\documentclass[sigconf]{acmart}
\usepackage{xcolor}

\usepackage{multirow}

\AtBeginDocument{%
  \providecommand\BibTeX{{%
    \normalfont B\kern-0.5em{\scshape i\kern-0.25em b}\kern-0.8em\TeX}}}

\begin{document}

\title{A Survey on the Robustness of Feature Importance and Counterfactual Explanations}

\author{Saumitra Mishra}
\affiliation{%
  \institution{J.P. Morgan AI Research}
  \city{London}
  \country{UK}
}
\email{saumitra.mishra@jpmorgan.com}


  \author{Sanghamitra Dutta}
  \affiliation{%
   \institution{University of Maryland}
   \city{College Park}
   \country{USA}
  }
\email{sanghamd@umd.edu}

  \author{Jason Long}
  \affiliation{%
   \institution{J.P. Morgan AI Research}
   \city{London}
   \country{UK}
  }
\email{jason.x.long@jpmorgan.com}

  \author{Daniele Magazzeni}
  \affiliation{%
   \institution{J.P. Morgan AI Research}
   \city{London}
   \country{UK}
  }
\email{daniele.magazzeni@jpmorgan.com}






\renewcommand{\shortauthors}{Mishra, et al.}

\begin{abstract}

There exist several methods that aim to address the crucial task of understanding the behaviour of AI/ML models. Arguably, the most popular among them are local explanations that focus on investigating model behaviour for individual instances. Several methods have been proposed for local analysis, but relatively lesser effort has gone into understanding if the explanations are robust and accurately reflect the behaviour of underlying models. In this work, we present a survey of the works that analysed the robustness of two classes of local explanations (feature importance and counterfactual explanations) that are popularly used in analysing AI/ML models in finance. The survey aims to unify existing definitions of robustness, introduces a taxonomy to classify different robustness approaches, and discusses some interesting results. Finally, the survey introduces some pointers about extending current robustness analysis approaches so as to identify reliable explainability methods.

\end{abstract}





\maketitle

\section{Introduction}


Machine learning (ML) models are being increasingly relied on to make decisions in a wide variety of contexts. Sometimes these decisions can have serious repercussions for individuals, with many examples to be found in the health, legal or financial sectors~(e.g., ~\cite{Angwin_web_2016}). In such circumstances, explanations for a model's output can help engender trust in the decision or help individuals take actions in order to obtain a more desirable outcome later. 

Explanations for machine learning models fall into several broad categories~\cite{Gilpin_dsaa_2018}, and are often evaluated along a number of distinct metrics~\cite{Bhatt_ijcai_2020}. In this survey we will touch on two key classes of explainability techniques, feature importance methods and counterfactual explanation methods, while focusing on one particular metric: the robustness of explanations. In XAI, the term `robustness' may refer to a number of related but distinct measures of how much explanations for a machine learning system may change under certain restricted changes to that system. Examples include the sensitivity of explanation techniques to choices (such as hyperparameters) which may be made `behind the scenes', or the effects of small changes to the input point or the model on explanations provided by the techniques.



Robustness has been argued as an important desideratum of local explanation methods~\cite{Hancox-Li_fat_2020}. However, in the literature there exists a lack of agreement about the meaning of robustness in the context of XAI. Moreover, researchers have used different terms to refer to the idea of robust explanations. For example,~\cite{Yang_arXiv_2019} used the term `sensitivity' to define how explanations change to changes in inputs (or models), but~\cite{Melis_arXiv_2018} used the term `robustness' for a similar idea. Similarly, some researchers have used the term `stability' to refer to how sensitive an explanation method is to its hyperparameters~\cite{Warnecke_arXiv_2020}.

In this survey, we try to unify the existing definitions (and terminologies) under the umbrella term `robustness'. We define an explanation method as robust if the explanations it generates remain same (or similar) under certain scenarios. Consider, for example, 
the context of actionable recourse~\cite{Karimi_arXiv_2020}, where the recipient of an adverse decision (e.g., credit denial) made by a machine learning model wishes to understand what actions they might take to change their outcome. There is a temporal element to this problem that plays an important role: in the time that it may take an individual to act on an explanation, other input features or the model itself may have changed in some limited way. If the explanations provided to individuals are very sensitive in response to such minor changes then the user of the explanation may find that the actions that they select ultimately fail to secure the changes that they desire.

Another key area in which the robustness of explanations can play a central role is in the context of assessing a model's ability to generalise. Explainability methods can often be used to determine whether a model's decisions are `right for the right reasons', and hence whether the model will remain accurate when faced with unseen data~\cite{Sturm_ieeetmm_2014}. Since this new data may have a slightly different distribution to previous data, explanations lacking in robustness may obscure the similarities in model behaviour and make it more difficult to trust the model's transferability.

In the next section, we will give a more technical introduction to the different varieties of robustness that have been introduced in the literature, and some examples where these definitions are applicable. We will then discuss robustness in the context of feature importance methods and separately in counterfactual explanations.

\section{Taxonomy of robustness analysis}
\label{sec:sec3}


In this section, we introduce the scenarios that have been used to analyse the robustness of explainability methods. Moreover, we use the scenarios to categorise the current methods for robustness analysis into three categories discussed below. Importantly, we focused only on works that analysed the robustness of feature importance and counterfactual explanations methods.

\begin{itemize}
    \item \textbf{Robustness to input perturbations:} This scenario involves keeping the machine learning model unchanged and analysing the behaviour of explainability methods to slight perturbations to model inputs \cite{Kindermans_arXiv_2017, Melis_arXiv_2018, Dombrowski_neurips_2019, Ghorbani_aaai_2019,slack2021counterfactual,dominguez2022adversarial}. Such input perturbations could be introduced deliberately by an adversary or could result from changes in data distribution. Importantly, all the perturbation approaches ensure that model predictions for the perturbed inputs are same (or quite similar) as model predictions for the original inputs. 
    
    \item \textbf{Robustness to model changes:} This scenario involves modifying the underlying machine learning model, but keeping the model inputs unchanged \cite{Adebayo_neurips_2018, Heo_neurips_2019, Dimanov_aaai_ws_2020, Anders_arXiv_2020, Slack_aies_2020,rawal2020can,upadhyay2021towards,dutta2022robust,black_consistent_2021}. Importantly, the methods in this scenario ensure that the new model has similar predictive performance as the original model. Similar to the above category, changes in an ML model could happen as a part of ML production cycle (e.g., due to changes in data distribution, addition of new training data, using a new type of model) or could be induced by an adversary.


    \item \textbf{Robustness to hyperparameters selection:} This scenario involves examining the sensitivity of explanations to changes in the hyperparameters of an explainability method~\cite{Zhang_arXiv_2019, Mishra_ijcnn_2020, Bansal_cvpr_2020}. Such an analysis is important as often using explainability methods requires deciding about several hyperparameters (e.g., the number of perturbed samples in LIME~\cite{Ribeiro_kdd_2016}, baseline in integrated gradients~\cite{Sundarrajan_icml_2017}), but the absence of ground-truth to verify explanations and the lack of clear guidelines for selecting hyperparameters makes the process non-trivial.

\end{itemize}

\subsection{Feature importance methods}


We now use the taxonomy from above to discuss works that analysed the robustness of a class of post-hoc local explainability methods called feature importance (attribution) methods. The goal of these methods is to assign an importance score to an input feature indicating the influence of the feature in the model prediction for the input. Formally, given a model $M$ and an input $x \in \mathcal{R}^n$, the feature importance methods output an attribution vector $A \in \mathcal{R}^n$, where $A^i$ denotes the importance of $i^{th}$ feature in the prediction $M(x)$. Importantly, the importance scores may have different meanings depending on whether the explainability method performs sensitivity analysis or function decomposition~\cite{Montavon_dsp_2018}.

\textbf{Robustness to input perturbations:} Within this category the majority of works focused on analysing the robustness of gradient-based saliency maps that are specific to analysing neural network models (differentiable models). For example,~\citet{Kindermans_arXiv_2017} demonstrated that perturbing inputs by simply adding a constant shift causes several gradient-based saliency methods to attribute incorrectly. Others designed novel objective functions to demonstrate that most of the popular saliency methods can be forced to generate arbitrary explanations and attributed this to certain geometrical properties of neural networks (e.g., shape of decision boundary)~\cite{Ghorbani_aaai_2019, Dombrowski_neurips_2019}.~ \citet{Melis_arXiv_2018} added to the results from above by demonstrating that not only model-agnostic methods (LIME~\cite{Ribeiro_kdd_2016}, SHAP~\cite{Lundberg_neurips_2017}) are also non-robust, but also they are considerably more non-robust than the gradient-based saliency methods. They introduced local Lipshitz continuity as a metric to quantify robustness for local input perturbations.

\textbf{Robustness to model changes:} Similar to the previous category, most approaches focused on analysing methods in the context of neural network models. Moreover, most of the approaches focused on learning an adversarial model with predictive performance similar to the  original one, but that can be used to generate manipulated explanations. For example,~\citet{Heo_neurips_2019} designed an objective function to demonstrate that some saliency methods are non-robust to adversarial model manipulations.~\citet{Anders_arXiv_2020} extended the previous work to show that it is always possible to find an adversarial model that has exactly the same behaviour as the original model but with manipulated explanations. Other works extended the analysis to show that even model-agnostic methods are non-robust~\cite{Dimanov_aaai_ws_2020, Slack_aies_2020}. Importantly, this category of robustness is of particular relevance to finance as a bad actor during an audit can easily present a model with similar accuracy as original, but with explanations that hide the bias contained in the original model. In another direction,~\citet{Adebayo_neurips_2018} performed model manipulation by replacing a neural network with its randomly initialised version with same architecture and understanding if the saliency maps change for the randomly initialised model. Interestingly, they found that some methods always fail this test.

\textbf{Robustness to hyperparameter selection:} This last category of robustness analysis is comparatively less explored. 
Researchers have demonstrated across multiple domains and model types that LIME has multiple sources of uncertainty in its explanations~\cite{Mishra_ijcnn_2020, Zhang_arXiv_2019} and it is important to select hyperparameters carefully to generate meaningful explanations.~\citet{Bansal_cvpr_2020} demonstrated for deep image classification models that not only LIME, but also several gradient-based explanation methods are significantly sensitive to hyperparameters (e.g., random seed).

\vspace{-1mm}
\subsection{Counterfactual explanations}


Here we discuss a new and emerging class of local explanations that go beyond feature attribution, i.e., counterfactual explanations~\cite{verma2020counterfactual,wachter2017counterfactual}. Consider a scenario where a machine learning model denies loan to an applicant (a data point $x\in \mathcal{R}^n$). Given such a data point, the broad goal of counterfactual explanations is to provide suggestions to the applicant on what changes they can possibly make to achieve their desired outcome, e.g., increase income by $10K$ to be approved for the loan. This is usually attained by finding one (or more) relevant data points for which the model produces the desired outcome.  Existing literature examines a variety of desiderata on such counterfactual explanations, e.g., proximity to $x$ to ensure smallest possible change for the applicant, or change in as few features as possible, or changes that are more ``realistic'' either because they lie on the data manifold or adhere to the underlying causal assumptions. For a more detailed survey of different existing approaches to generate counterfactual explanations, we refer the reader to \cite{verma2020counterfactual, Karimi_arXiv_2020} and the references therein.

However, counterfactual explanations might also be unreliable in several situations. Here, we survey some recent works that address the unreliabilities in counterfactual explanations.

\textbf{Robustness to input perturbations:} In \cite{slack2021counterfactual}, the authors show how one might employ adversarial training to achieve drastically different counterfactual explanations under a small perturbation to the input. They focus on existing techniques of generating counterfactual explanations that rely on ``hill-climbing'' (i.e., gradient-based approaches that find counterfactuals by solving an optimization). To understand this concept, let us first understand how gradient-based approaches to find counterfactuals would typically work.

Given a data-point $x\in \mathcal{R}^n$ with undesired outcome and a model $f(\cdot)$, one can find a data-point with desired outcome as follows:
\begin{equation}
    x'=\arg \min (f(x')-1)^2 + d(x,x'). \label{eq:cf_finding}
\end{equation}
Here, $d(x,x')$ denotes the distance between the points $x$ and $x'$, that captures the effort required to change $x$ to $x'$ (could be $l_1$ loss, $l_2$ loss, or pair-wise distance metrics, e.g.,~\cite{verma2020counterfactual,upadhyay2021towards}). We assume that the desired outcome of the model is $1$. Typically, such an optimization is solved using gradient descent until a local minima is reached as long as $f(\cdot)$ is differentiable. Now consider two data points $x$ and $x+\delta$ that are quite close to each other, e.g., two loan applicants who differ just a little in their income. One might want the counterfactual explanations for these two applicants to be similar. However, \cite{slack2021counterfactual} introduces an adversarial objective for training $f(\cdot)$ that can lead to drastically different counterfactual explanations for these two data points, one of which may involve significantly more effort (higher $d(x,x')$). This happens because the optimization in \eqref{eq:cf_finding} converges to two different local minima (due to the adversarial training of $f(\cdot)$). Furthermore, such a technique could be used to embed unfairness in counterfactual explanations across groups, e.g., with respect to sensitive attributes such as gender, race, etc. It remains an interesting challenge to analyze and understand how one can effectively audit and detect such unfairness embedded into model training.

\textbf{Robustness to model changes:} 
Another aspect of unreliability arises when the existing model is updated (possibly due to retraining on the same data or newly acquired data). In such situations, it is often desirable that the counterfactual explanations that have already been provided to applicants in the past still remain valid. For instance, consider an applicant who was denied loan, and the counterfactual explanation provided to the applicant was to increase their income by $10K$. Now suppose that they indeed increase their income by $10K$ and reapply for the loan. If the model is no longer the same, there is no guarantee that their loan will now be approved, leading to potential mistrust and liability concerns for the counterfactual explanations. Thus, another challenge is to generate counterfactual explanations that already account for such potential changes to the model in future (at least within certain limitations). During the generation of counterfactual explanations, one might want to find data points that not only lie on the other side of the decision boundary, but are also expected to lie on the other side after certain anticipated changes to the model.

Changes to the model could happen due to various reasons. For instance, one might want to retrain the model on the same data but with different hyperparameters, or switch to a different model altogether, e.g., from linear regression to neural networks. For such changes to the model, counterfactual explanations that strictly lie on the data manifold, (e.g., using causality-inspired approaches \cite{konig2021causal} or other observational approaches~\cite{poyiadzi2020face}, etc.) might be more robust than counterfactuals that lie outside the data manifold. 

Often, the data on which the model is retrained might also be changed, e.g., due to corrections, temporal or geo-spacial distribution shifts, etc. In \cite{rawal2020can}, the authors analyze how the probability of invalidation of counterfactual explanations are affected by such changes on the underlying data distribution. Table~\ref{tab:taxonomy} summarises the discussed robustness scenarios. 

\textbf{Robustness to hyperparameter selection:} There is less research directly into the robustness of counterfactual explanations to changes to the hyperparameters of the explanation method. However, in many works the counterfactual search problem is reduced to a loss minimisation problem in which several competing objectives (such as differing output, distance from query, diversity, etc) are weighted against one another~\cite{verma2020counterfactual, Karimi_arXiv_2020}. Balancing the terms in such a loss function can be challenging, and different weights can change the loss landscape, and hence the explanations, dramatically~\cite{multiobjective}.


\begin{table*}
  \caption{A summary of robustness analysis scenarios for two types of post-hoc local explainability methods - feature importance and counterfactuals. NN, RF, GBT, LR, MN-NB, SVM refer to neural network, random forest, gradient booted trees, logistic regression, multinomial naive Bayes and support vector machine, respectively. m-agnostic and m-dependent refer to the domain of the explainability methods, where m refers to model.}
  \label{tab:taxonomy}
  \begin{tabular}{ccccccc}
    \toprule
    Reference & \multicolumn{3}{c}{ Explanation Methods} & Robustness Scenario & Models & Data Types\\
    & type & m-agnostic & m-dependent & & \\
    \midrule
    \cite{Kindermans_arXiv_2017} & feature importance & no & yes & input perturbation & NN & images \\
    \cite{Melis_arXiv_2018} & feature importance & yes & yes & input perturbation & RF, NN & tabular, images \\
    \cite{Ghorbani_aaai_2019} & feature importance & no & yes & input perturbation & NN & images \\
    \cite{Dombrowski_neurips_2019} & feature importance & no & yes & input perturbation & NN & images \\

    \cite{Adebayo_neurips_2018} & feature importance & no & yes & model manipulation & NN & images\\
    \cite{Heo_neurips_2019} & feature importance & no & yes  & model manipulation & NN & images\\
    \cite{Anders_arXiv_2020} & feature importance & no & yes & model manipulation & LR, NN & tabular, images \\
    \cite{Slack_aies_2020} & feature importance & yes & no & model manipulation & RF & tabular\\
    \cite{Dimanov_aaai_ws_2020} & feature importance & yes & yes & model manipulation & NN & tabular\\

    \cite{Zhang_arXiv_2019} & feature importance & yes & no & hyperparameters selection & RF, GBT, MN-NB & tabular, text\\
    \cite{Bansal_cvpr_2020} & feature importance & yes & yes & hyperparameters selection & NN & images \\
    \cite{Mishra_ijcnn_2020} & feature importance & yes & no & hyperparameters selection & NN & audio \\
    
    
    
    \cite{slack2021counterfactual} & counterfactuals & yes & yes & input perturbation & NN & tabular\\
    
    \cite{rawal2020can} & counterfactuals & no & yes & model manipulation & LR, RF, GBT, SVM, NN & tabular\\
    
  \bottomrule
\end{tabular}
\end{table*}

\vspace{-2.5 mm}
\section{Robust Explanations}

As discussed earlier, empirical and theoretical analysis demonstrated that the majority of popular feature importance and counterfactual explanation methods are non-robust. Recently, there have been some works that aim to tackle this challenge. 

In the context of feature importance methods,~\cite{Anders_arXiv_2020, Dombrowski_pr_2022} have proposed approaches to make gradient-based methods for DNNs significantly more robust.~\citet{Anders_arXiv_2020} took inspiration from the field of manifold learning and proposed to project explanations along tangential directions of the data manifold.~\citet{Dombrowski_pr_2022} proposed three ways to improve the robustness of DNN explanations - (1) by training DNNs with weight decay; (2) by training using smoothed activation functions; and (3) by adding a regulariser for model's curvature in the training process. Similarly,~\citet{Lakkaraju_icml_2020} proposed the RObust Posthoc Explanations (ROPE) framework that generates explanations robust to changes in individual input (adversarial robustness) and changes in data distribution (distributional roustness). The authors demonstrate that without compromising on local fidelity the explanations generated by ROPE are more robust than LIME and SHAP. 

Robustness of counterfactual explanations is an interesting and challenging problem space. In \cite{dominguez2022adversarial}, the authors examine the problem of finding counterfactuals that are robust to small perturbations to the input under a causal model, and propose techniques to incorporate robustness during the counterfactual search. Ensuring robustness of counterfactuals under model changes is also another interesting direction. When the data is unaltered but the model changes due to retraining, counterfactuals that lie on the data manifold are more likely to remain valid than those that lie outside the manifold. Existing literature proposes several techniques to generate counterfactuals that lie on the data manifold (e.g.~\cite{konig2021causal,poyiadzi2020face,pawelczyk2020learning} among others). For a detailed survey of such techniques that find counterfactuals on the data manifold, we refer the reader to \cite{verma2020counterfactual} and the references therein.

When the underlying data distribution also changes, the problem becomes more challenging, and is largely unexplored with the notable exception of some recent works \cite{upadhyay2021towards,black_consistent_2021,dutta2022robust}. In  \cite{upadhyay2021towards}, the authors proposes an algorithm called ROAR that finds counterfactuals for models that would also remain valid if the model changes in the parameter space within certain bounds. To achieve this, \cite{upadhyay2021towards} uses a joint maximization-minimization-based approach during the search for counterfactuals for robustness. At each step of the iteration, the worst-case perturbation to the model is obtained (maximization), and then, an attempt is made to find counterfactuals for this model (minimization). In \cite{black_consistent_2021}, an alternate two-step technique is proposed for finding robust counterfactuals for deep neural networks that can be applied on top of existing techniques of counterfactual generation. After a counterfactual is found using an existing technique, \cite{black_consistent_2021} searches in its neighborhood for a more robust counterfactual using a stability criterion (that is defined for differentiable models). Both \cite{upadhyay2021towards} and \cite{black_consistent_2021} are able to find counterfactuals that are quite robust to model changes with some increase in the distance to the original point ($L_1$ or $L_2$ cost).

Tree-based models often pose additional challenges in counterfactual generation due to their non-differentiable nature. In \cite{dutta2022robust}, a model-agnostic criterion for stability is introduced that attempts to quantify how robust a counterfactual is going to be to model changes under retraining, and comes with some desirable theoretical properties.  Given a counterfactual, the stability criterion considers the model outputs for a bunch of points around it (Gaussian distributed): it promotes counterfactuals with high mean output value, and penalizes high variability in the output value. The proposed strategy RobX~\cite{dutta2022robust} works with any counterfactual generation
method (base method) and searches for robust
counterfactuals by iteratively refining the counterfactual generated by the base method using the proposed metric Counterfactual Stability. The generated counterfactuals are not only robust, but also lie on the data manifold (in terms of the metric Local Outlier Factor) with some increase in the distance to the original point ($L_1$ or $L_2$ cost).

\section{Discussion}

In this section, we briefly introduce some pointers for future research. For example, robustness analysis with input perturbations has been mostly aimed towards deliberate adversarial modifications to input instances, which one may argue may not happen in the real world. However, a simpler and real-world analysis could involve identifying neighbouring points to an instance from the data set and analysing if a method generates similar explanations for them. Similarly, it would be interesting to analyse if the non-robustness of ML (especially deep learning) models to targeted input perturbations is related to the lack of robustness for gradient-based explanations.

Importantly, most of the robustness analysis has been done for differentiable models and large-scale image data sets. However, in the context of finance, the use of deep neural networks is still not prevalent and tree-based models have been a popular substitute~\cite{Bracke_boe_2019}. Thus, detailed analysis of model-agnostic explanation methods in the context of large-scale financial data sets would be needed to understand the extent of non-robustness of explanations in the context of finance. Moreover, in the context of finance, more study is required to understand the impact of hyperparameters choice on the performance of LIME and SHAP.

In the context of counterfactual explanations, the analysis of the impact of hyperparameters on the robustness of a method has largely been an unexplored area. However, such a study is important as the recent trend in developing novel counterfactual methods involves adding more (loss) terms to the objective function with less discussion about how to weight each term.

\vspace{-0.25 cm}

\section{Conclusion}

In this paper, we presented a review of existing approaches to analyse the robustness of two popular classes of post-hoc local explainability methods - feature importance and counterfactual explanations. We defined robustness in the context of explainability and introduced a taxonomy to categorise existing approaches into three categories. Specifically, we categorise the present approaches (feature importance and counterfactual explanations) into - methods that perturb inputs, methods that manipulate ML models, and methods that change hyperparameters values. We discussed several methods from each category and highlighted their key results. Overall, the majority of explainability methods are non-robust and hence employing them to understand models used in safety-critical applications is risky. We also reviewed some recent methods that propose approaches to tackle some of the robustness challenges discussed above. Finally, we presented some prospective research directions to further analyse explainability methods, especially in the context of finance.

In this work, we focused on one criterion for analysing explainability methods. However, there exist works that analysed explainability methods using other important criteria~\cite{Bhatt_ijcai_2020, Jacovi_acl_2020} (e.g., fidelity, efficiency, usefulness, complexity). There is an urgent need for a large-scale benchmarking of popular explanability methods. Some recent efforts have noted this both for feature importance~\cite{Warnecke_arXiv_2020, Yang_arXiv_2019} and counterfactual methods~\cite{Pawelczyk_arXiv_2021}. However, we would need more such efforts to ensure we can trust the explanation methods and their explanations.



\bibliographystyle{ACM-Reference-Format}
\bibliography{robust-refer}

\end{document}